\documentclass[acmsmall,screen,nonacm]{acmart}
\setcopyright{none}

\usepackage{listings}
\usepackage{setspace}
\usepackage{xcolor}
\usepackage{booktabs}
\usepackage{stmaryrd}

\usepackage{etoolbox}
\usepackage{needspace}
\BeforeBeginEnvironment{lstlisting}{\Needspace*{6\baselineskip}}

\usepackage{letltxmacro}
\LetLtxMacro\lstinlineorig\lstinline
\renewcommand{\lstinline}[1][]{\lstinlineorig[basicstyle=\ttfamily,#1]}

\newcommand{\ie}{\textit{i.e.},}
\newcommand{\etc}{\textit{etc.}}

\newcommand{\untop}{\mathcal{U}}
\newcommand{\abstop}{\top}
\newcommand{\absbot}{\bot}
\newcommand{\atom}{\alpha}
\renewcommand{\not}[1]{\neg #1}
\newcommand{\cnj}[2]{#1 \wedge #2}
\newcommand{\dsj}[2]{#1 \vee #2}
\newcommand{\imp}[2]{#1 \to #2}
\newcommand{\nxt}[1]{\mathcal{X}#1}
\newcommand{\unt}[2]{#1 \untop #2}
\newcommand{\evt}[1]{\mathcal{F}#1}
\newcommand{\alw}[1]{\mathcal{G}#1}
\newcommand{\trace}{\tau}
\newcommand{\hole}{\cdot}
\newcommand{\interp}[1]{\llbracket #1 \rrbracket}

\newif\ifanon\anonfalse		
\newcommand{\inh}[1]{\textcolor{black!55}{#1}}

\author{Konstantinos Kogkalidis}
\title{Backward through Time, Algebraically}

\begin{document}

\begin{abstract}
Linear temporal logic is a modal extension of propositional logic that allows one to state how a system should behave over time.
Its canonical domain is the booleans, but discretely-valued judgements are of little use in steering softly-valued systems (neural policies, adaptive controllers, sequence models, \etc{}).
In such cases, the goal formula's (dis)satisfaction becomes a training signal, and differentiability becomes a prime concern.
Candidate differentiable semantics abound, but navigating them is tricky.
Implementations, where available, are shallow embeddings, demanding an upfront commitment to a single semantic algebra and its (usually implicit) conduct.
The paper casts the reader as a functional programmer asked to come to terms with this predicament, and refusing.
Out of that refusal comes an evaluation engine that is algebra-generic and amenable to differentiation, together with an executable specification of the algebras it can accept.
Various algebras are implemented and audited for their behavior, both forward and backward.
Each algebra turns out to be a choice of which direction to disappoint, and how.
Everything described (and more) is part of the PyTorch library \texttt{telos}, to be found at \url{https://github.com/konstantinosKokos/telos}.
\end{abstract}

\maketitle

\section{Syntax}
You stumble upon a logical formula, and a mostly innocuous one at that; it contains only familiar propositional connectives, plus a few funny symbols.
Some vague but undeniable authority tasks you with evaluating the formula as it wades through \textit{time}.
You do not flinch.
The funny symbols attain their denotation as the problem makes its way into your memory.
It is old, and all but solved.
You dust off the authoritative scriptures of old~\cite{pnueli1977temporal} and recall the syntax of \textit{linear temporal logic}:
\begin{equation*}
	\phi, \psi :=
			\abstop \
			| \ \absbot \
			| \ \atom \
			| \ \not{\phi} \
			| \ \cnj{\phi}{\psi} \
			| \ \dsj{\phi}{\psi} \
			| \ \imp{\phi}{\psi} \
			| \ \nxt{\phi} \
			| \ \unt{\phi}{\psi}
\end{equation*}
Of the two non-classical operators, one makes \textit{instant} sense; $\nxt{\phi}$ (read: \textit{next} $\phi$) is just $\phi$, one tick ahead of whenever \textit{now} is.
The other takes a few moments \textit{longer}; $\unt{\phi}{\psi}$ (read: $\phi$ \textit{until} $\psi$) turns out to be two promises uttered at once: that a moment favorable to $\psi$ exists, and that $\phi$ carries the burden of truth every tick of the way, arrival included.
One symbol is \textit{a step}; the other is \textit{a search, within a search}.
Two more symbols sometimes travel in the same circles: $\evt{\phi}$ (read: \textit{finally} $\phi$), and $\alw{\phi}$ (read: \textit{globally} $\phi$).
Both are just $\untop$ in disguise.
The former is the search relieved of its side condition, $\evt{\phi} := \unt{\abstop}{\phi}$; the latter denies a future where $\phi$ lapses, $\alw{\phi} := \not{\evt{\not{\phi}}}$.
The two find use in specifying temporal conduct: \textit{safety}, demanding that nothing bad ever happens, and \textit{liveness}, that something good eventually does~\cite{lamport1977proving}.

On its own, the formula is only half an instruction.
The other half is a \textit{trace} $\trace$, the world the formula speaks of.
The trace is a two-dimensional assignment; it says what the value of each atom $\atom$ is at each tick $t$, over a finite time window of $T$ ticks.
You pair the two together and get the full instruction, a \textit{judgement} $\trace \models \phi$.
You roll up your sleeves and, honestly, it doesn't seem like much of a task; pattern match on $\phi$, step the trace to the right on $\nxt{\hole}$, sweep from the far end for $\unt{\hole}{\hole}$, \etc

Just as you announce your compute-readiness, the data comes in.
It is not what you expected.
For starters, the trace matrix is not boolean.
It contains numbers of some \lstinline|dtype=float32|, and carries a flag that reads \lstinline|requires_grad=True|; a PyTorch~\cite{paszke2019pytorch} Tensor.
You flinch.
The numbers, you learn, are freshly squeezed out of a neural network, and that flag is the umbilical cord that keeps them attached to the machinery that produced them.
Whoever sent them your way does not much care what the formula's verdict is right now; they intend to \textit{make} it favorable by repeatedly nudging the trace's producer, for as long as that takes.
You relent; you are picky and precious about your instruments, but concessions are sometimes necessary.
The grammar ports in an hour's work, sugar included, your ADTs now emulated via abstract class inheritance.
Your first semantic instinct is to round: call anything above one half \lstinline|True|, and the scriptures are back in business.
The rounding restores your truth-values, and it costs only everything.
The step function's derivative flickers between zero and infinity; whatever loss is built downstream of your verdict cannot find its way through the rounding and back into the network.
Your mistake was brandishing a verdict (\textit{whether} the formula is satisfied) when you were asked for feedback (\textit{how} satisfied it is); feedback that must additionally be \textit{differentiable} in every number the trace contains.
The voice did not speak of bools; it spoke of \textit{values}, writ large.
You do your due diligence perusing the bazaar of fuzzy logics~\cite{klement2000triangular}.
You get dizzy; there's a few options too many, and you're not invested enough to make the call yourself.
You shrug; \textit{``I'll just make this algebra-generic and let the user pick, I guess''}.

\section{Semantics, Abstract}\label{sec:abstract}
You gather and superimpose the go-to algebras and blank out the places where they diverge.
The impression left underneath is a blueprint of their shared abstraction:
a \textit{carrier} with two \textit{distinguished elements}, standing in for $\abstop$ and $\absbot$,
and a pointwise operation over that carrier for each logical connective.
You don't let backprop obligations dissuade you from purity and peace of mind; PyTorch can deal with the effectful book-keeping on its own.
You start writing%
\footnote{Read \lstinline|Fn| as shorthand for the unutterable \lstinline|Callable| annotation. Read \lstinline|T| as shorthand for \lstinline|torch.Tensor|. Consider syntax errors a deliberate attempt at condensation (most of them are).}:
\begin{lstlisting}
class Algebra(abc.ABC, torch.nn.Module):
    top: T; bot: T
    def meet(self, x: T, y: T) -> T: ...
    def join(self, x: T, y: T) -> T: ...
    def impl(self, x: T, y: T) -> T: ...
    def neg(self, x: T) -> T: ...
\end{lstlisting}
The names are borrowed from lattice theory on credit; whether an algebra earns them is a question momentarily deferred.
But what of \textit{time}?
The funny symbols are still semantically adrift.
Anchoring them takes little effort; you know well that the ``temporal'' part is just first-order quantification (\ie{} iteration) in disguise.
Time enters the play as the last tensor axis, and every temporal need reduces to sweeping some operation along it:
\begin{lstlisting}
def scan(fn: Fn[[T, T], T]) -> Fn[[T], T]:
    def f(x: T) -> T:
    	# unbind splits off the time axis; read: stack . scanl1 fn . unbind
        return torch.stack(list(accumulate(x.unbind(-1), func=fn)), dim=-1)
    return f

def fold(fn: Fn[[T, T], T], initial: T) -> Fn[[T], T]:
    def f(x: T) -> T:
    	# read: foldl fn initial . unbind
        return reduce(fn, x.unbind(-1), initial)
    return f

def span(fn: Fn[[T], T], neutral: T, bot: T) -> Fn[[T], T]:
    def f(x: T) -> T:
        n = x.size(-1)
        # an upper triangular (n, n) bool matrix
        mask = torch.triu(torch.ones(n, n, device=x.device)).bool()
        # unsqueeze tiles the signal into n rows, and where is if then else in tensor form;
        # so row t keeps ticks from t onward, earlier ones blanked to fn's neutral element
        rows = torch.where(mask, x.unsqueeze(-2), neutral)
        # now fn sweeps every window x[t..t'], and sub-diagonal cells read bot
        return torch.where(mask, fn(rows), bot)
    return f
\end{lstlisting}
Armed with these, your algebra-concocting recipe is finally complete:
\begin{lstlisting}
class Algebra(abc.ABC, torch.nn.Module):
    ...
    def running_meet(self, x: T) -> T: return scan(self.meet)(x)
    def running_join(self, x: T) -> T: return scan(self.join)(x)
    def forall(self, x: T) -> T: return fold(self.meet, self.top)(x)
    def exists(self, x: T) -> T: return fold(self.join, self.bot)(x)
    def span_meet(self, x: T) -> T: return span(self.running_meet, self.top, self.bot)(x)
\end{lstlisting}
Note what has quietly transpired: the derived operations are spelled out once, in terms of the four abstract primitives and not much else.
You will lean on this harder than it may now appear; the definitions above are less of a default implementation and more of a \textit{specification} that just so happens to be runnable.

\section{Interface}

The sleeves, rolled up for a while now, finally meet their pattern match.
Evaluation is structural recursion over the formula, each case deferring to the algebra:
\begin{lstlisting}
class Model(torch.nn.Module):
    def forward(self, A: Algebra, judgement: Judgement) -> T:
        (trace, phi) = judgement
        def go(phi: Formula) -> T:
            match phi:
                case Variable(x):
                    return trace[x]
                case Negation(Until(AbstractTop(), Negation(x))):  # globally, desugared
                    return A.running_meet(go(x).flip(-1)).flip(-1)
                case Negation(x):
                    return A.neg(go(x))
                case Conjunction(l, r):
                    return A.meet(go(l), go(r))
                ...
                case Next(x):
                     return torch.nn.functional.pad(go(x)[..., 1:], (0, 1), value=A.bot)
                case Until(AbstractTop(), r):  # finally, desugared
                    return A.running_join(go(r).flip(-1)).flip(-1)
                case Until(l, r):
                    # span_meet: cell (t, t') holds l's meet over the segment t..t'
                    # the unsqueeze broadcasts r across rows: window, meet witness
                    # exists sweeps the t' axis, leaving one verdict per t
                    return A.exists(A.meet(A.span_meet(go(l)), go(r).unsqueeze(-2)))
        return go(phi)[..., 0]
\end{lstlisting}
Writing $\interp{\hole}$ to denote the evaluator \lstinline|go| under some algebra and trace,
$\nxt{\hole}$ pads its unoccupied final tick with $\interp{\absbot}$; a certain pessimism about the end of time that you \textit{elect} to own.
The two aliases of $\untop$ are amenable to compute-friendly searches over suffixes, \ie{} running reductions over reversed time; \textit{finally} and \textit{globally} then sit only a primitive apart.
The more general $\unt{\hole}{\hole}$ reads exactly as promised: every window of $\phi$ is pitted against its witness $\psi$, and the best outcome is kept.
The evaluator is thus complete, blissfully oblivious of algebra peculiarities and \lstinline|dtype| commitments; it shall interpret every formula under every algebra ever written (including the ones yet to be).

\section{Semantics, Concrete}

Before celebrating generality, you reconcile with the scriptures through a quick regression test.
To no one's surprise, boolean semantics are trivial to recover:
\begin{lstlisting}
class Boolean(Algebra):
    top, bot = torch.tensor(True), torch.tensor(False)
    def meet(self, x, y): return x & y
    def join(self, x, y): return x | y
    def impl(self, x, y): return ~x | y
    def neg(self, x): return ~x
\end{lstlisting}
Instantiate, evaluate and the answers come back as promised.
Whatever the abstraction may end up costing, it won't be correctness; the scriptures have just been demoted to a special case of the blueprint.
Confident, you return to the fuzzy bazaar, freshly relieved of the burden of choice.
You begin transcribing; first a helper:
\begin{lstlisting}
class FuzzyBase(Algebra, abc.ABC):
    top, bot = torch.tensor(1.), torch.tensor(0.)
    def neg(self, x): return self.top - x
\end{lstlisting}
Then, the probabilist's default:
\begin{lstlisting}
class Product(FuzzyBase):
    def meet(self, x, y): return x * y
    def join(self, x, y): return x + y - x * y
    def impl(self, x, y): return torch.where(x == self.bot, self.top, (y / x).clamp(max=1))
\end{lstlisting}
The evaluator obliges; every judgement now returns a number that quacks like a probability.

Back to the deferred credit check; the names \lstinline|meet| and \lstinline|join| came with lattice-theoretic obligations.
Associativity, commutativity and monotonicity you wave through, and involution and the De Morgan dualities with them; all hold universally in this corner of the world, a fact audited and asserted.
The other properties of interest are recited below:
\begin{equation*}
	\begin{tabular*}{\linewidth}{@{\extracolsep{\fill}}cccc@{}}
	\textit{idempotence} & \textit{absorption} & \textit{distributivity} & \textit{complementation} \\[2pt]
	$\interp{\cnj{\phi}{\phi}} = \interp{\phi}$ & $\interp{\cnj{\phi}{(\dsj{\phi}{\psi})}} = \interp{\phi}$ & $\interp{\cnj{\phi}{(\dsj{\psi}{\xi})}} = \interp{\dsj{(\cnj{\phi}{\psi})}{(\cnj{\phi}{\xi})}}$ & $\interp{\cnj{\phi}{\not{\phi}}} = \interp{\absbot}$ \\
	$\interp{\dsj{\phi}{\phi}} = \interp{\phi}$ & $\interp{\dsj{\phi}{(\cnj{\phi}{\psi})}} = \interp{\phi}$ & $\interp{\dsj{\phi}{(\cnj{\psi}{\xi})}} = \interp{\cnj{(\dsj{\phi}{\psi})}{(\dsj{\phi}{\xi})}}$ & $\interp{\dsj{\phi}{\not{\phi}}} = \interp{\abstop}$
	\end{tabular*}
\end{equation*}

\lstinline|Product|'s audit is brief, and unflattering.
Idempotence immediately bounces back: $x * x$ agrees with $x$ nowhere but at the domain's edges.
A formula $\phi$ pinned at $x$ gets $\interp{\alw{\phi}} = x^T$; the value of \textit{always} is now tied to its duration.
Absorption and distributivity fail on the same grounds, with more bureaucracy and no insights to show for it.
Complementation fares no better: $x * (1-x)$ hovers consistently above $\interp{\absbot}$, while the middle, supposedly excluded, is in fact perfectly included.
Product, in short, has earned almost none of its borrowed names.
No matter; in assuming no law, the evaluator couldn't have broken any.
You record the audit all the same, and give in to your propensity for automation.
Each law is a function of the function(s) it constrains:
\begin{lstlisting}
def idempotent(op: Fn[[T, T], T]) -> Fn[[T], bool]:
    def f(x: T) -> bool: return torch.allclose(op(x, x), x)
    return f

def absorption(meet: Fn[[T, T], T], join: Fn[[T, T], T]) -> Fn[[T, T], bool]:
    def f(x: T, y: T) -> bool:
        return torch.allclose(meet(x, join(x, y)), x) & torch.allclose(join(x, meet(x, y)), x)
    return f
\end{lstlisting}
Strict equality makes way for lax negotiation, an artifact of finite precision arithmetic.
Auditing becomes a guided search for counter-examples: a loop of every law against every algebra, over a batch of (initially) random tensors~\cite{claessen2000quickcheck,maciver2019hypothesis}.

Tests in place, the rest of the wares transcribe just as quickly, and audit just as unevenly (Table~\ref{tab:zoo}).
\begin{table}[h]
	\caption{Common algebras, audited. \textit{Diff.} marks algebras differentiable in their inputs; \textit{Train.} marks parametric families whose parameter is itself a learnable tensor. Lettered entries hold only as indicated by the corresponding note. Law entries are checked mechanically.}
	\label{tab:zoo}
	\small
	\begin{tabular}{llcccccc}
	\toprule
	Algebra & Carrier & Diff. & Train. & Idem. & Abs. & Dist. & Comp. \\
	\midrule
	\lstinline|Boolean|        & $\mathbb{B}$              	&                              &            & \checkmark & \checkmark & \checkmark & \checkmark 	\\
	\lstinline|Goedel|         & $[0,1]$                   	& $^{a}$             			&            & \checkmark & \checkmark & \checkmark &            	\\
	\lstinline|KleeneDienes|   & $[0,1]$                   	& \checkmark                   &            & \checkmark & \checkmark & \checkmark &            	\\
	\lstinline|Lukasiewicz|    & $[0,1]$                   	& \checkmark                   &            &            &            &            & \checkmark 	\\
	\lstinline|Product|        & $[0,1]$                   	& \checkmark                   &            &            &            &            &            	\\
	\lstinline|Robustness|     & $\mathbb{R}\cup\{\pm\infty\}$& \checkmark               		&            & \checkmark & \checkmark & \checkmark &            	\\
	\lstinline|Frank|          & $[0,1]$                   	& \checkmark                   & \checkmark & $^{b}$     & $^{b}$     & $^{b}$     & $^{c}$     	\\
	\lstinline|Hamacher|       & $[0,1]$                   	& \checkmark                   & \checkmark &            &            &            &            	\\
	\lstinline|Yager|          & $[0,1]$                   	& \checkmark                   & \checkmark & $^{c}$     & $^{c}$     & $^{c}$     & $^{d}$     	\\
	\lstinline|SchweizerSklar| & $[0,1]$                   	& \checkmark                   & \checkmark &            &            &            & $^{e}$     	\\
	\lstinline|AczelAlsina| 	& $[0,1]$              			& \checkmark                	& \checkmark & $^{c}$     & $^{c}$     & $^{c}$     &            	\\
	\lstinline|Dombi|    		& $[0,1]$               		& \checkmark               		& \checkmark & $^{c}$     & $^{c}$     & $^{c}$     &				\\
	\lstinline|SugenoWeber|    & $[0,1]$                   	& \checkmark                   & \checkmark &            &            &            & $^{b}$     	\\
	\lstinline|LSE|            & $\mathbb{R}\cup\{\pm\infty\}$ & \checkmark               	& \checkmark & $^{c}$     & $^{c}$     & $^{c}$     &            	\\	\bottomrule
	\end{tabular}

	\smallskip\smallskip
	\begin{minipage}{.88\textwidth}
	{\footnotesize
	$^{a}$\,\lstinline|impl| is not differentiable in its first argument. \quad
	$^{b}$\,as $p \to 0$. \quad
	$^{c}$\,as $p \to \infty$. \quad
	$^{d}$\,at $p = 1$. \quad
	$^{e}$\,for $p \geq 1$.
	}
	\end{minipage}
\end{table}
\lstinline|Boolean| sits at the top; the only lattice proper.
Its descendants below trade lawfulness for the spoils of differentiability.
On \{0, 1\}-valued traces, the boundary conditions leave zero freedom: every algebra on the unit interval agrees with \lstinline|Boolean| exactly (modulo the necessary type coercion).
It is in the interior where variation lives.
\lstinline|Robustness| and \lstinline|LSE| stand apart from the rest, having wandered off the unit interval altogether.
These are imported from \textit{signal temporal logic}~\cite{maler2004monitoring,fainekos2009robustness,leung2023backpropagation}; their carrier is the extended real line and their negation plain arithmetic: $\interp{\neg{\phi}} = -\interp{\phi}$.
The lower half of the table hosts the generalized, parametric families~\cite{klement2000triangular}: each carries a parameter $p$ declared as a learnable tensor, so that the algebras themselves (not just the traces they judge) can be subject to numerical optimization.
Varying $p$ smoothly deforms an algebra, up to the limit where it meets a law-abiding neighbor: Frank, Acz\'el-Alsina, Dombi and Yager all harden to G\"odel, and LSE to Robustness.
Looking back, you gleefully note how the abstract machinery has been doing double duty as a reference manual, and its instantiations now serve as literature review.

\section{Semantics, Faster}

Alas, utility stands in opposition to generality.
As traces outgrow politely-sized examples, the evaluator buckles under unfavorable asymptotes: spans quadratic in memory, folds and scans linear in depth, each tick a round-trip through the interpreter.
Correct and general, sure. But also heavy and slow.

The memory footprint seems impossible to tackle.
But what of time?
The offense committed needs some clarifying; it's not the \textit{work} that's at fault.
Each of the $T$ ticks has to be read once, and each is indeed read exactly once.
The culprit is the serial nature of \textit{depth} iteration: not the number of steps taken, but the number of steps spent in wait.
Each fold treads like a procession, single file across a device built millions of lanes wide.
Mid-diagnosis, you pause and instinctively shift your attention towards the audit table and the laws unceremoniously inscribed therein.
Your mind's eye rests on a column loudly absent: \textit{associativity}.
For the logician, a license to re-bracket; for the engineer, a license to start tomorrow's work today~\cite{blelloch1989scans}.
A fold free to regroup needs not proceed one tick at a time.
What it regroups \textit{into}, though, is not for associativity to decide; the closed form, when one exists, depends on the rest of the algebra's properties.

\lstinline|Goedel|, for instance, is idempotent; its meet is an extremum, and running extrema are long canonized builtins:
\begin{lstlisting}
class Goedel(FuzzyBase):
    ...
    def running_meet(self, x): return torch.cummin(x, dim=-1).values
    def running_join(self, x): return torch.cummax(x, dim=-1).values
\end{lstlisting}
The override here is less of an over-rule and more of a claim of agreement with the fold it displaces.
Ordinarily one would owe a proof; here, the displaced fold could (and still \textit{can}) run, so the claim simply joins the audit as one more law, searched for counter-examples alongside the rest.
The two sides of overriding appeal to different sensibilities: one is optimization; the other is stating tiny little theorems that also happen to execute.

\lstinline|Product|, despite holding no special law, draws the same luck; its running meet is a cumulative product; kernel enough to also be found on the shelf.
The shelf stocks nothing for the running join, but doesn't really need to; De Morgan's law transforms a join into a negated meet over negated parts:
\begin{lstlisting}
class Product(FuzzyBase):
    ...
    def running_meet(self, x): return torch.cumprod(x, dim=-1)
    def running_join(self, x): return self.neg(self.running_meet(self.neg(x)))
\end{lstlisting}
Suspiciously, \lstinline|LSE| finds its kernel waiting on the same shelf, almost as if commissioned~\cite{pant2017smooth,gilpin2021smooth}:
\begin{lstlisting}
class LSE(Algebra):
    ...
    def running_join(self, x): return torch.logcumsumexp(self.p * x, dim=-1) / self.p
\end{lstlisting}
The parametric families hold no idempotence, and no kernel awaits them by name.
Their meets, though, look eerily alike:
\begin{lstlisting}
 ... return torch.clamp(1 - ((1 - x)**p + (1 - y)**p)**(1/p), min=0.)  # Yager
 ... return torch.exp(-((-torch.log(x))**p + (-torch.log(y))**p)**(1/p))  # Aczel-Alsina
 ... return 1 / (1 + (((1 - x)/x)**p + ((1 - y)/y)**p)**(1/p))  # Dombi
\end{lstlisting}
You squint a bit, and the nested complexity begins to look less like variety and more like camouflage.
You reprise the opening trick: superimpose them, blank out the places where they diverge:
\begin{lstlisting}
 ... return g_inv(g(x) + g(y))  # the impression left underneath
\end{lstlisting}
What is left behind this time is \textit{addition}, albeit in different coordinates.
Each meet factors as $g^{-1}(g(x) + g(y))$ for some strictly decreasing \textit{generator} $g$ with $g(1) = 0$~\cite{ling1965representation}; where a family's meet saturates at $0$, its pseudo-inverse clamps to match.
In generator coordinates the fold is a running sum; \lstinline|torch.cumsum| has been there for you all along:
\begin{lstlisting}
class Archimedean(Algebra, ABC):
    def g(self, x: T) -> T: ...
    def g_inv(self, x: T) -> T: ...

    def meet(self, x, y): return self.g_inv(self.g(x) + self.g(y))
    def impl(self, x, y): return torch.where(x <= y, self.top, self.g_inv(self.g(y) - self.g(x)))
    def running_meet(self, x): return self.g_inv(torch.cumsum(self.g(x), dim=-1))
    def forall(self, x): return self.g_inv(self.g(x).sum(dim=-1))
\end{lstlisting}
The joins mirror through negation, courtesy of De Morgan duality, and the whole parametric wing gets to exercise its re-bracketing license.
In hindsight, the move retroactively explains the shelf's earlier generosity: a cumulative product is a \lstinline|cumsum| in $-\log$ coordinates, and \lstinline|logcumsumexp| carries its generator loudly in its name.
\lstinline|Product| now quietly re-parents, surrendering its hand-written joins to the family it secretly belonged to all along.
One line of the listing was not asked for: \lstinline|impl| computes the \textit{residuum}, the largest value whose meet with $x$ stays at or below $y$; it fell out of the coordinates unbidden, and it retires a hand-written implication in every family it touches.
Instantiating is now embarrassingly brief:
\begin{lstlisting}
class Lukasiewicz(Archimedean, FuzzyBase):
    def g(self, x): return 1 - x
    def g_inv(self, s): return torch.clamp(1 - s, min=0.)
\end{lstlisting}
That is really the entire algebra; two lines of coordinates, everything else derived.

The licenses reach $\unt{\hole}{\hole}$ as well.
The general case wants a meet over every window $x_{t..t'}$; there are $T^2$ of those, and computing each on its own is how one arrives, honestly and carefully, at a cubic algorithm.
But windows sharing a start differ only in their end; each is a prefix of the next, and a single running sweep along a row delivers the whole family at once.
This is the triangular construction \lstinline|span| has been performing all along, for which no special license is needed.
What associativity buys is the sweep itself: handed a closed-form \lstinline|running_meet|, \lstinline|span_meet| collapses into one kernel launch over one $T \times T$ tensor.

Who, then, still keeps the fold?
In the current catalogue, no one.
It now stands only where it first appeared: as a specification and a fallback, held in reserve for the algebras the evaluator promised to interpret, the ones yet to be written.
With time tamed, it is tempting to call the matter closed.

\section{Semantics, Backward?}
You yield to the temptation.
The setup is the minimal workable one: a formula encoding some preference, a sweep over algebras, an optimizable trace with the appropriate carrier.
The goal is simple: evaluate the formula, turn the value into a loss term, backprop to the trace, update it towards satisfaction.
Nothing breaks; the loss even descends\dots{} at a geological timescale.
Suspicious, you put a single backward pass under the loupe.
The feedback is differentiable, as requested; it is also rather empty.

Emptiness takes three forms.
The first is \textit{selection}.
\lstinline|Goedel| and \lstinline|Robustness| rely on extrema, and their derivatives are indicator functions~\cite{varnai2020robustness}.
However long the trace or involved the formula, the backward pass singles out one tick of one atom, which then bears the computation's entire credit; every other entry receives an exact zero.
The second is \textit{saturation}.
\lstinline|Lukasiewicz| relies on repeated clamping: on traces of any serious length, outputs plateau and gradients are erased.
The third is \textit{decay}.
\lstinline|Product| distributes feedback credit among participants, across both time and (syntactic) space.
Regrettably, distribution does not counteract scarcity.
Chains of multiplications in the unit interval push values towards $\interp{\absbot}$, and collapsing values beget collapsing gradients.
This suspiciously tidy taxonomy happens to be a theorem~\cite{mostert1957structure,ling1965representation}.
Loosely, it claims that every continuous t-norm is an \textit{ordinal sum}, glued together from two archetypes, a \textit{nilpotent} one and a \textit{strict} one, with an \textit{idempotent} one at the seams; saturation, decay, and selection, in that order.
The classification is exhaustive; there is no fourth alternative to look for.

Two problems compound.
One you have just witnessed: standard fuzzy operators make for notoriously bad neural networks in their own right~\cite{krieken2022analyzing}.
The other is that naively unrolling a contracting operator yields a gateless and parameterless recurrent neural network, invariably condemned to vanishing gradients (even for well-behaved operators)~\cite{hochreiter1998vanishing}.
The moment time was declared quantification in disguise, both problems were committed to at once, and their effects seem to now be putting this enterprise to an unrewarding end.
The realization stings, then condemns: were all the concessions made in vain?

\section{Semantics, Lifted}
You anxiously search the wreckage for survivors.
On reread, the theorem's domain is continuous t-norms, which neither \lstinline|Robustness| nor \lstinline|LSE| are (their carrier being the extended real line).
Its selective behavior has burned the former; the latter was condemned by association alone, and without proper trial.
You put its backward pass under the same loupe, and find it\dots{} dense with gradients!
The per-tick shares of credit are softmax weights: they sum to one, and none of them is a zero.
A $\phi$ pinned to almost truth for $T$ ticks sees its feedback thinning as $1/T$ (\ie{} harmonically rather than exponentially); duration may dilute the credit, but it never extinguishes it.
Read against the taxonomy, the profile appears a threefold antidote: it is dense, so nothing is selected out; it is smooth, so nothing saturates; and it is normalized, so nothing decays.
A panacea, however, it is not: the audit's old complaint stands, and the verdict of \textit{always} is still tied to its duration.
Hold $\phi$ to finite truth, and every confirming tick drags $\interp{\alw{\phi}}$ lower; enough ticks, and the verdict-carrying sign flips to unsafety.
Conversely, $\interp{\evt{\phi}}$ climbs up with each tick spent not finding $\phi$, eventually reporting liveness where there is none.
The catalogue thus splits clean in two: sound verdicts with dead credit, or live credit with unsound verdicts; no algebra presently holds both halves.

But what makes \lstinline|LSE| so special?
You return to the theorem and give it another read, this time looking for hypotheses rather than verdicts: continuity, associativity, and the unit interval.
\lstinline|LSE| kept the first two and broke the third.
But then again, as the generator already confessed, \lstinline|LSE| is just \lstinline|Product| under bent coordinates.
The theorem has not been escaped; transported across the algebra isomorphism, it still classifies the survivor as strict.
And then it clicks: the theorem is indifferent to coordinate bending, but the backward pass is not; it's the unit interval doing the extinguishing all along!
The click resolves the dead credit question, but the unsound verdict problem still stands.
What's missing is idempotence: the sum nested within \lstinline|LSE| grows with its duration, and no coordinate bending can make it forget its length.

The two problems translate into two demands.
The credit must keep the only profile to have survived inspection: the shares must be the softmax.
The verdict must forget its duration: the sum must be normalized like a mean, since a mean of equals is an equal, indifferent to how many steps it averages over.
For once, credit assignment can be prescribed rather than discovered.
Both demands seemingly point at the same object: the first names the softmax outright, and the softmax is itself a normalized weighting scheme, so averaging under it is a mean by construction.
The emerging candidate is actually folklore: the \textit{Boltzmann} average, a mean whose weights are the softmax of its inputs.

But the demands exceed the spec, and transcription stalls at the first reduction.
The spec assumed no law, so that it could not break any.
Yet it did assume a shape: it derived quantification as the fold of a binary operator, native to the algebra's carrier.
The newcomer holds neither of the licenses that shape has been spending.
Its bracketings disagree: an average of averages is not the average of the whole, so a binary fold is a silent commitment to one of many bracketings~\cite{czogala1984associative}.
The fallback would still run it, only to correctly compute the wrong thing.
Furthermore, it has no mute member: every finite value carries weight in a weighted average, leaving nothing to fill \lstinline|span|'s blanks with.
For the first time, it is not an algebra that fails an audit, but the spec that fails an algebra.

How does one binarize an irreducibly variadic operator?
The usual trick of superimposing no longer applies; there are no two items to compare.
Then again, the trick's last run left a residue: the \lstinline|Archimedean| family.
It prescribes the recipe $g^{-1} \circ \mathrm{cumsum} \circ\, g$; a fold flanked by coordinate transformations.
You squint again, harder this time, and realize your perspective may have been somewhat myopic.
The fold never ran in the carrier; it ran in another space, picked exactly so that the fold behaves.
Reducibility is not a property of the operator alone; it is granted to the operator by the space it folds in.
The misattribution is excusable; the target space has so far been hiding under two independent coincidences.
One, the crossing cost nothing semantically, $g$ being invertible.
And two, it cost nothing syntactically, $g$ mapping \lstinline|T| to \lstinline|T|.
Breaking away from both, you consider the minimum workable properties of the target space: associativity (to license speed), identity (to license padding), and\dots{} that's it.
You lift the operator to a \textit{monoidal} state that just so happens to contain tensors in some shape or form:
\begin{lstlisting}
class State(ABC):
    duration: int; device: torch.device
    def combine(self, other: Self) -> Self: ...
    def neutral(self) -> Self: ...
    def zip_with(self, other: Self, fn: Fn[[Tensor, Tensor], Tensor]) -> Self: ...
\end{lstlisting}
The licenses are spent immediately, on what will soon become \lstinline|scan|'s faster replacement~\cite{hillis1986data} in \lstinline|State|-space:
\begin{lstlisting}
def sweep[S: State](states: S) -> S:
    def shifted(acc: S, k: int) -> S:
        def delay(s: Tensor, n: Tensor) -> Tensor:
            # s, arriving k ticks late; the neutral n covers the vacancy
            return torch.cat([n.expand(*s.shape[:-1], k), s[..., :-k]], dim=-1)
        return acc.zip_with(states.neutral(), delay)
    ks = takewhile(lambda k: k < states.duration, (1 << j for j in count()))
    return reduce(lambda acc, k: shifted(acc, k).combine(acc), ks, states)
\end{lstlisting}
Read temporally, a \lstinline|State| is a \textit{summary} of an interval; \lstinline|combine| merges adjacent intervals without revisiting their ticks: a monitor's memory that splices as readily as it extends.
\lstinline|sweep| is that monitor run over every prefix at once.
The same licenses admit the triangular tiling of \lstinline|span|, its blanks this time filled by the monoid's unit:
\begin{lstlisting}
def windows[S: State](states: S) -> S:
    n = states.duration
    mask = torch.triu(torch.ones(n, n, device=states.device)).bool()
    def tile(s: Tensor, i: Tensor) -> Tensor:
        # as in span: row t keeps ticks from t onward; the blanks now read neutral
        return torch.where(mask, s[..., None, :], i)
    return sweep(states.zip_with(states.neutral(), tile))
\end{lstlisting}

The abstraction meets its first inhabitant, whose state is shaped nothing like its value:
\begin{lstlisting}
class BoltzmannState(State):
    max: Tensor; weight: Tensor; wsum: Tensor

    def combine(self, other: BoltzmannState) -> BoltzmannState:
        m = torch.maximum(self.max, other.max)
        wa = torch.exp(self.max - m)
        wb = torch.exp(other.max - m)
        return BoltzmannState(
            max=m,
            weight=wa * self.weight + wb * other.weight,
            wsum=wa * self.wsum + wb * other.wsum,
        )

    def neutral(self) -> BoltzmannState:
        return BoltzmannState(
            max=self.max.new_tensor(float('-inf')),
            weight=self.weight.new_tensor(0.),
            wsum=self.wsum.new_tensor(0.),
        )
\end{lstlisting}
The state carries the weighted mean in three pieces: a numerator, a denominator, and a running maximum~\cite{milakov2018online,dao2022flashattention}.

The blueprint, in its current form, is not ready to consume the new arrivals.
The upside is that \lstinline|State|s are transient and internal; they never escape the confines of a computation, so they can be tucked away by an \lstinline|Algebra| that only ever deals with \lstinline|Tensor|s.
The downside is that \lstinline|Algebra| spelled out reductions as folds of binary primitives, folds which subclasses inherit as both a fallback and a specification; this now becomes a liability.
The solution is a retroactive split: \lstinline|Algebra| keeps the signatures and surrenders the bodies; the runnable spec relocates, virtually identical, into \lstinline|Folded(Algebra, ABC)|, and the entire catalogue reparents to it.
A new sibling then emerges:
\begin{lstlisting}
class Lifted[S: State](Algebra, ABC):
    def embed(self, x: T) -> S: ...
    def readout(self, s: S) -> T: ...

    def meet(self, x, y): return self.readout(self.embed(x).combine(self.embed(y)))
    def join(self, x, y): return self.neg(self.meet(self.neg(x), self.neg(y)))
    def impl(self, x, y): return self.join(self.neg(x), y)
    def running_meet(self, x): return self.readout(sweep(self.embed(x)))
    def running_join(self, x): return self.neg(self.running_meet(self.neg(x)))
    def forall(self, x): return self.running_meet(x)[..., -1]
    def exists(self, x): return self.running_join(x)[..., -1]

    def span_meet(self, x: T) -> T:
        n = x.size(-1)
        mask = torch.triu(torch.ones(n, n, device=x.device)).bool()
        return torch.where(mask, self.readout(windows(self.embed(x))), self.bot)
\end{lstlisting}
Each running reduction factors as $\mathrm{readout} \circ \mathrm{sweep} \circ \mathrm{embed}$, with $\mathrm{sweep}$ determined by \lstinline|combine| alone; the crossing is the factorization's only free parameter.
The span is the old triangular construction transported across the crossing.
Meet and join are distinct aggregations over a shared monoid, so they could only ever differ in their crossings; rather than implement a second pair, the abstraction assumes involutive negation so that join can mirror the meet through negation, just like the \lstinline|Archimedean| family.
The second implicit assumption is that each embed is a section of the readout, $\mathrm{readout} \circ \mathrm{embed} = \mathrm{id}$ so that each value sent can come back unharmed (the mean of one thing is the thing).
The converse is not true; the readout is free to forget, and what is forgotten is forever lost.
And that is what the imprint of non-associativity really is: two different brackets forget different things, at different places.
Both assumptions then turn to laws, and join the audit alongside the rest.
The newcomer gets a name and a listing, made of one crossing pair and a negation:
\begin{lstlisting}
class Boltzmann(Lifted[BoltzmannState]):
    b: T

    def embed(self, x): return BoltzmannState(-self.b * x, torch.ones_like(x), x)
    def neg(self, x): return -x
    def readout(self, s): return s.wsum / s.weight
\end{lstlisting}
\lstinline|Boltzmann| runs at the catalogue's full speed, its credit dense by construction, but its laws still owed.
Once more, it is tempting to call the matter closed.

\section{Semantics, Backward!}

This time around you know better, and jump straight to the loupe.
Initially, the backward pass invites optimism; the feedback is differentiable, as requested, and this time it is anything but empty: dense, smooth, and normalized, not a tick uncredited.
The audit interrupts: \textit{monotonicity}, previously universal, now fails.
The disproving exhibit is compact and unquestionable: a two-tick judgement of $\evt{\phi}$, one strong peak and one faint tick; \textit{raise} the faint tick and the verdict \textit{drops}.
It takes a second, but the cause becomes obvious in hindsight: the weights are elected by the very values under judgement, so a straggler promoted gains weight (while still testifying low!), and the average obliges, downward.

The mistake is a rehash of your very first one, except this time in reverse: the softmax belonged to the feedback, but was spent instead on the verdict.
The demands said where the softmax must show, not where to install it; the shares reside in the backward pass, and the survivor had been holding them there all along.
The repair has to be ``backwards''-engineered from the survivor: keep its aggregation, and let its sum too become a mean.
The constraints converge: at fixed duration, division becomes subtraction by a constant, and constants are invisible to the backward pass; the survivor's shares carry over verbatim.
The resulting operator already has a name, \textit{mellowmax}, and a reputation as a better-behaved alternative to the Boltzmann average~\cite{asadi2017alternative}.
It is also a quasi-arithmetic mean, \ie{} the averaging sibling of the \lstinline|Archimedean| family, its exponential tilting a generator in disguise~\cite{kolmogorov1930moyenne,nagumo1930klasse}.
In \lstinline|State|-space, \lstinline|Boltzmann|'s successor requires a single substitution: the weighted sum retires, and in its place sits a \textit{counter}, given a field of its own.
\begin{lstlisting}
class MellowmaxState(State):
    max: Tensor; weight: Tensor; count: Tensor

    def combine(self, other: MellowmaxState) -> MellowmaxState:
        m = torch.maximum(self.max, other.max)
        wa = torch.exp(self.max - m)
        wb = torch.exp(other.max - m)
        return MellowmaxState(
            max=m,
            weight=wa * self.weight + wb * other.weight,
            count=self.count + other.count,
        )

    def neutral(self) -> MellowmaxState:
        return MellowmaxState(
            max=self.max.new_tensor(float('-inf')),
            weight=self.weight.new_tensor(0.),
            count=self.count.new_tensor(0.),
        )
\end{lstlisting}
The algebra follows suit:
\begin{lstlisting}
class Mellowmax(Lifted[MellowmaxState]):
    b: T

    def embed(self, x): return MellowmaxState(-self.b * x, torch.ones_like(x), torch.ones_like(x))
    def neg(self, x): return -x
    def readout(self, s): return (s.max + torch.log(s.weight) - torch.log(s.count)) / -self.b
\end{lstlisting}

Almost mechanically, you reach for another backward pass.
Not unlike before, the feedback is dense, smooth and normalized.
Unlike before, it is also non-negative, exactly matching \lstinline|LSE|'s profile: the softmax, verbatim.
As designed, the verdict does forget its duration; the mean is idempotent, so $\phi$ pinned at $x$ reads $\interp{\alw{\phi}} = x$, at any $T$.
No more selection, no more saturation, and no more decay%
\footnote{At finite $\beta$. As $\beta$ grows, the shares collapse towards one-hotness and mellowmax hardens into \lstinline|Robustness|. Selection was never banished, only pushed to a boundary, and the distance to the boundary is the parameter.}.
As is customary, the audit interrupts, on a debt thought settled: the \textit{unit}.
A mean has no mute member; for any $\abstop$ large enough to approximate infinity, conjunction with truth pushes the conjunct further towards truthhood.
The unit's loss is unavoidable; a mean divides by the count of its voters, and a unit is an abstainer, a value not to be counted.
The loss is not, however, insurmountable; with $\abstop$ read as a distinguished boundary value, a numerical inequality suffices to map it to the monoid's unit:
\begin{lstlisting}
class Mellowmax(Lifted[MellowmaxState]):
    ...

    def embed(self, x):
        member = x < self.top  # top abstains
        return MellowmaxState(
            max=torch.where(member, -self.b * x, -torch.inf),
            weight=member.float(), count=member.float()
        )

    def readout(self, s):
        vote = (s.max + torch.log(s.weight) - torch.log(s.count)) / -self.b
        return torch.where(s.count > 0, vote, self.top)  # an empty ballot reads top
\end{lstlisting}
Despite the algebra not being unital in the carrier, the unit can be emulated in \lstinline|State|-space, which is where computation happens anyway; a half-win.
The audit concludes without further surprises.
Monotonicity, involution, De Morgan duality, and the section law all pass, as expected.
Also as expected, associativity, the residuum and the lattice honorifics (absorption, distributivity, complementation) all fail, and implication retracts to its material form.
The search concludes: duration-invariant verdicts and live credit, at last held by a single algebra.

\section{Semantics, in Hindsight}
The matter can finally be called closed.
The concessions were indeed worth it; more than that, every concession can now be understood as a structural trespass, a property broken in the verdict world for a behavior gained in the feedback world.
None of the ruckus was perceived by the evaluator, which has kept its promise throughout: every formula, under every algebra.
The result is a bunch of algebras for the user to pick, in varying grades of well-behavedness, properly qualified, and with at least one of them working; they all get to meet one another under Table~\ref{tab:zoo2}.
You walked into the bazaar a dizzy customer; you walk out its keeper; the dizziness has been cured by careful appraisal rather than choice.
Satisfied with the artifact, you start recounting your turmoils into a somewhat exaggerated second-person narrative, which you reckon could make for a good enough functional pearl.
In time, you will get both feedback and a verdict; this time around, neither will be of your own design.

\begin{table}[h]
	\caption{%
	Table~\ref{tab:zoo} revisited: the extended catalogue and audit suite.
	\inh{Greyed} cells are carried over unchanged from Table~\ref{tab:zoo}; black cells are new here.
	Law entries are still checked mechanically.
	\textit{Credit} is not a law but a classification, forward-determined and backward-observed.}
	\label{tab:zoo2}
	\smaller[2]
	\setlength{\tabcolsep}{4.5pt}
	\begin{tabular}{llcccccccccc}
	\toprule
		Algebra & Carrier & Diff. & Train. & Assoc. & Idem. & Abs. & Dist. & Comp. & Unital. & Mono. & Credit \\
	\midrule
	\lstinline|Boolean|        & \inh{$\mathbb{B}$}                  &                 &                 & \checkmark & \inh{\checkmark} & \inh{\checkmark} & \inh{\checkmark} & \inh{\checkmark} & \checkmark & \checkmark & n/a            \\
	\lstinline|Goedel|         & \inh{$[0,1]$}                       & \inh{$^{a}$}    &                 & \checkmark & \inh{\checkmark} & \inh{\checkmark} & \inh{\checkmark} &                  & \checkmark & \checkmark & \textit{sel.}  \\
	\lstinline|KleeneDienes|   & \inh{$[0,1]$}                       & \inh{\checkmark}&                 & \checkmark & \inh{\checkmark} & \inh{\checkmark} & \inh{\checkmark} &                  & \checkmark & \checkmark & \textit{sel.}  \\
	\lstinline|Robustness|     & \inh{$\mathbb{R}\cup\{\pm\infty\}$} & \inh{\checkmark}&                 & \checkmark & \inh{\checkmark} & \inh{\checkmark} & \inh{\checkmark} &                  & \checkmark & \checkmark & \textit{sel.}  \\
	\lstinline|Lukasiewicz|    & \inh{$[0,1]$}                       & \inh{\checkmark}&                 & \checkmark &                  &                  &                  & \inh{\checkmark} & \checkmark & \checkmark & \textit{sat.}  \\
	\lstinline|Yager|          & \inh{$[0,1]$}                       & \inh{\checkmark}& \inh{\checkmark}& \checkmark & \inh{$^{c}$}     & \inh{$^{c}$}     & \inh{$^{c}$}     & \inh{$^{d}$}     & \checkmark & \checkmark & \textit{sat.}  \\
	\lstinline|SchweizerSklar| & \inh{$[0,1]$}                       & \inh{\checkmark}& \inh{\checkmark}& \checkmark &                  &                  &                  & \inh{$^{e}$}     & \checkmark & \checkmark & \textit{sat.}  \\
	\lstinline|SugenoWeber|    & \inh{$[0,1]$}                       & \inh{\checkmark}& \inh{\checkmark}& \checkmark &                  &                  &                  & \inh{$^{b}$}     & \checkmark & \checkmark & \textit{sat.}  \\
	\lstinline|Product|        & \inh{$[0,1]$}                       & \inh{\checkmark}&                 & \checkmark &                  &                  &                  &                  & \checkmark & \checkmark & \textit{dec.}  \\
	\lstinline|Frank|          & \inh{$[0,1]$}                       & \inh{\checkmark}& \inh{\checkmark}& \checkmark & \inh{$^{b}$}     & \inh{$^{b}$}     & \inh{$^{b}$}     & \inh{$^{c}$}     & \checkmark & \checkmark & \textit{dec.}  \\
	\lstinline|Hamacher|       & \inh{$[0,1]$}                       & \inh{\checkmark}& \inh{\checkmark}& \checkmark &                  &                  &                  &                  & \checkmark & \checkmark & \textit{dec.}  \\
	\lstinline|AczelAlsina|    & \inh{$[0,1]$}                       & \inh{\checkmark}& \inh{\checkmark}& \checkmark & \inh{$^{c}$}     & \inh{$^{c}$}     & \inh{$^{c}$}     &                  & \checkmark & \checkmark & \textit{dec.}  \\
	\lstinline|Dombi|          & \inh{$[0,1]$}                       & \inh{\checkmark}& \inh{\checkmark}& \checkmark & \inh{$^{c}$}     & \inh{$^{c}$}     & \inh{$^{c}$}     &                  & \checkmark & \checkmark & \textit{dec.} \\
	\lstinline|LSE|            & \inh{$\mathbb{R}\cup\{\pm\infty\}$} & \inh{\checkmark}& \inh{\checkmark}& \checkmark & \inh{$^{c}$}     & \inh{$^{c}$}     & \inh{$^{c}$}     &                  & \checkmark & \checkmark & \textit{dense} \\
		\lstinline|Boltzmann|      & $\mathbb{R}\cup\{\pm\infty\}$   & \checkmark      & \checkmark      & $^{f}$     & \checkmark       &                  &                  &                  & $^{f}$     &            & \textit{dense} \\
	\lstinline|Mellowmax|      & $\mathbb{R}\cup\{\pm\infty\}$       & \checkmark      & \checkmark      & $^{f}$     & \checkmark       &                  &                  &                  & $^{f}$     & \checkmark & \textit{dense} \\
	\bottomrule
	\end{tabular}

	\smallskip\smallskip
	\begin{minipage}{.88\textwidth}
	{\footnotesize
	\inh{$^{a}$\,\lstinline|impl| is not differentiable in its first argument. \quad}
	\inh{$^{b}$\,as $p \to 0$. \quad}
	\inh{$^{c}$\,as $p \to \infty$. \quad}
	\inh{$^{d}$\,at $p = 1$. \quad}
	\inh{$^{e}$\,for $p \geq 1$. \\}
	$^{f}$\,in \lstinline|State|-space, up to readout; not in the carrier.
	}
	\end{minipage}
\end{table}

\bibliographystyle{ACM-Reference-Format}
\bibliography{references}

\end{document}